\documentclass[letterpaper]{article} 
\usepackage{aaai2026}

\usepackage{times}  
\usepackage{helvet}  
\usepackage{courier}  
\usepackage[hyphens]{url}  
\usepackage{graphicx} 
\usepackage{natbib}  
\usepackage{caption} 
\usepackage{multirow}
\usepackage{makecell}
\usepackage{booktabs}
\usepackage{amssymb}
\usepackage{algorithm}
\usepackage{algorithmic}
\usepackage{amsmath}

\DeclareMathOperator*{\argminB}{argmin}   
\usepackage{newfloat}
\usepackage{listings}
\DeclareCaptionStyle{ruled}{labelfont=normalfont,labelsep=colon,strut=off} 
\floatstyle{ruled}
\newfloat{listing}{tb}{lst}{}
\floatname{listing}{Listing}

\usepackage[x11names]{xcolor}                  
\definecolor{codecomment}{RGB}{96,160,160}
\lstdefinelanguage{PseudoPy}{
  keywords={for, in, if, else, return, def},
  morecomment=[l]{\#},
  sensitive=true,
}
\newcommand{\cmtmath}[1]{\textcolor{codecomment}{\(#1\)}}

\makeatletter
\@ifundefined{isChecklistMainFile}{
  \newif\ifreproStandalone
  \reproStandalonetrue
}{
  \newif\ifreproStandalone
  \reproStandalonefalse
}
\makeatother

\ifreproStandalone
\usepackage{times}
\usepackage{helvet}
\usepackage{courier}
\usepackage{xcolor}
\title{An Efficient Model-Driven Groupwise Approach for Atlas Construction}
\author{
  Ziwei Zou\textsuperscript{\rm 1,3,4},
  Bei Zou\textsuperscript{\rm 1},
  Xiaoyan Kui\textsuperscript{\rm 1{$\dagger$}},
  Wenqi Lu\textsuperscript{\rm 2},
  Haoran Dou\textsuperscript{\rm 3,4},
  Arezoo Zakeri\textsuperscript{\rm 3,4},\\
  Timothy Cootes\textsuperscript{\rm 3},
  Alejandro F Frangi\textsuperscript{\rm 3,4},
  Jinming Duan\textsuperscript{\rm 3,4{$\dagger$}}
}
\affiliations{
    \textsuperscript{\rm 1} School of Computer Science and Engineering, Central South University, China \\
    \textsuperscript{\rm 2} Department of Computing
and Mathematics, Manchester Metropolitan University, UK\\
    \textsuperscript{\rm 3} Division of Informatics, Imaging and Data Sciences, University of Manchester, UK \\
    \textsuperscript{\rm 4} Centre for Computational Imaging and Modelling in Medicine, University of Manchester, UK

}

\usepackage{bibentry}

\begin{document}

\maketitle

\begin{abstract}

Atlas construction is fundamental to medical image analysis, offering a standardized spatial reference for tasks such as population-level anatomical modeling. While data-driven registration methods have recently shown promise in pairwise settings, their reliance on large training datasets, limited generalizability, and lack of true inference phases in groupwise contexts hinder their practical use. In contrast, model-driven methods offer training-free, theoretically grounded, and data-efficient alternatives, though they often face scalability and optimization challenges when applied to large 3D datasets. In this work, we introduce DARC (Diffeomorphic Atlas Registration via Coordinate descent), a novel model-driven groupwise registration framework for atlas construction. DARC supports a broad range of image dissimilarity metrics and efficiently handles arbitrary numbers of 3D images without incurring GPU memory issues. Through a coordinate descent strategy and a centrality-enforcing activation function, DARC produces unbiased, diffeomorphic atlases with high anatomical fidelity. Beyond atlas construction, we demonstrate two key applications: (1) One-shot segmentation, where labels annotated only on the atlas are propagated to subjects via inverse deformations, outperforming state-of-the-art few-shot methods; and (2) shape synthesis, where new anatomical variants are generated by warping the atlas mesh using synthesized diffeomorphic deformation fields. Overall, DARC offers a flexible, generalizable, and resource-efficient framework for atlas construction and applications.

\end{abstract}


\section{Introduction}

An atlas is a representative medical image that encapsulates the average anatomical structure derived from a population of individual subjects. In medical image analysis, atlases serve as standardized spatial references, facilitating a range of downstream tasks such as automated segmentation, regional quantitative analysis, lesion detection, and population-level analysis. The construction of anatomical atlases typically relies on groupwise image registration, a process intrinsically linked to pairwise registration techniques.

In the context of pairwise image registration, deep learning–based data-driven methods have achieved registration accuracy on par with conventional model-driven approaches that rely on iterative optimization \cite{thorley2021nesterov,27,duan2023arbitrary, 55}. A key advantage of data-driven methods lies in their computational efficiency at inference time; once trained, these models can perform registration tasks significantly faster than iterative model-driven approaches \cite{balakrishnan2019voxelmorph, 27, 56}. However, this efficiency comes at the cost of several limitations. Data-driven registration frameworks typically require large amounts of training data to achieve robust performance \cite{57, 59}. The estimation of deformation or displacement fields via neural network regression is inherently susceptible to approximation errors, which may degrade registration accuracy, particularly in regions with complex anatomical variation \cite{26, 58}. Furthermore, the generalization ability of data-driven methods is often limited, especially when applied to data distributions that deviate from those seen during training. In contrast, model-driven approaches are training-free and therefore highly data-efficient. They optimize deformation or displacement fields directly through iterative algorithms, avoiding intermediate approximation errors introduced by neural networks. Additionally, model-driven methods are largely invariant to data distribution shifts, making them inherently more generalisable and readily applicable to new datasets or clinical settings. 

In the context of atlas construction via groupwise registration, data-driven approaches have not demonstrated substantial improvements in computational efficiency over conventional model-driven methods. While trained neural networks can deliver near-instantaneous inference in pairwise registration tasks, this advantage does not directly translate to groupwise settings. Recent data-driven groupwise registration frameworks commonly adopt what is referred to as \textit{neural-network-based optimization }\cite{12,21}, in which the network learns to construct a template image by optimizing its parameters solely during the training process \cite{13, 14, 15, 16}. In other words, these methods lack an inference phase. The learned network parameters must be retrained or fine-tuned to construct a new template when applied to a different group of images, or even the same group with a different number of images. This reliance on fine-tuning or retraining undermines the core advantage of data-driven methods, namely rapid inference. As a result, the utility of data-driven groupwise registration remains limited, particularly in applications that demand efficiency and generalizability across heterogeneous datasets. Although recent studies, such as \cite{22}, have proposed data-driven methods that generate atlases in a single forward pass without fine-tuning or retraining, these approaches are heuristic, often lack theoretical grounding, and may not generalize well across anatomical structures or imaging protocols. 

Model-driven atlas construction, therefore, remains an active research area. These methods are training-free, theoretically interpretable, and generally robust across different image groups. However, optimization can be challenging, particularly when incorporating complex constraints such as centrality and diffeomorphisms. In addition, computing atlases from large-scale datasets remains computationally demanding, as the GPU memory footprint can grow exponentially when processing a number of high-resolution 3D volumetric images simultaneously. To address these challenges, we present a model-driven groupwise registration approach for atlas construction and introduce a novel coordinate descent algorithm to optimize it. We refer to this approach as DARC (\textbf{D}iffeomorphic \textbf{A}tlas \textbf{R}egistration via \textbf{C}oordinate descent) which has three main novelties:
\begin{itemize}
\item DARC incorporates a general dissimilarity data term that supports MSE, $L_1$, NCC, SSIM, or in theory any metric that measures the distance between the atlas image and each warped input image. For MSE and $L_1$, we show that closed-form solutions are available for updating the atlas image. For NCC and SSIM, we propose using stochastic gradient descent to perform the update. This design enables DARC to efficiently handle an arbitrary number of 3D images without hitting GPU memory limits. Additionally, we introduce an activation function to effectively enforce the zero-displacement constraint (centrality), ensuring that the resulting atlas image is unbiased with respect to any individual input image. DARC is implemented in PyTorch, enabling GPU acceleration and automatic differentiation, which makes it both simple and efficient. DARC is compared with state-of-the-art model- and data-driven approaches \cite{33, 34, 12, 16, 21}, demonstrating its superior performance. In addition, the results from DARC enable two novel applications described below.
\item We leverage the constructed atlas from DARC for one-shot segmentation. In this approach, only the atlas image requires annotation, which is then propagated to each subject through inverse deformation fields.  Our one-shot segmentation approach outperforms existing segmentation methods \cite{23, 24, 25}, owing to the accurate atlas construction achieved by DARC. 
\item Building on the atlas mesh and the resulting displacement fields from DARC, we propose a novel approach that implicitly generates a new virtual population of anatomies by synthesizing deformation fields and applying them to the atlas mesh. This approach ensures that the generated shapes naturally preserve point-wise correspondence, facilitating population-level shape analysis. In addition, the underlying deformation remains diffeomorphic, thereby avoiding mesh folding and maintaining anatomical plausibility. Thanks to the  alignment achieved by DARC, we demonstrate that shape synthesis using linear PCA models can perform on par with advanced diffusion models.
\end{itemize}

\section{Related Work}

\noindent 
\textbf{Model-driven methods} based on iterative optimization have long been the classical approach to atlas construction \cite{50}. Early groupwise registration methods focused on deformation plausibility and topological preservation, employing techniques such as iterative closest point algorithms for rigid and affine alignment \cite{35}, multiscale B-spline approaches \cite{51}, and the Large Deformation Diffeomorphic Metric Mapping (LDDMM) algorithm \cite{37}. Building on LDDMM, various extensions have been proposed, including unbiased atlas construction for large deformations \cite{2,3}, Riemannian geometric-median atlases \cite{5}, multimodal Bayesian infant brain atlases \cite{4}, EM-based atlas estimation \cite{38}, and PDE-constrained LDDMM \cite{39}. \cite{34} later introduced Symmetric Normalization (ANTs), the first symmetric, unbiased diffeomorphic atlas construction method. Another direction is hierarchical groupwise registration \cite{40,33,41,42,43}, which leverages local manifold structures to improve scalability, robustness, and efficiency. However, model-driven methods are difficult to optimize, especially with complex constraints or dissimilarity measures like NCC, and struggle to scale to large datasets without closed-form loss functions.


\begin{figure*}[htbp]
	\centering
	\includegraphics[width=\linewidth]{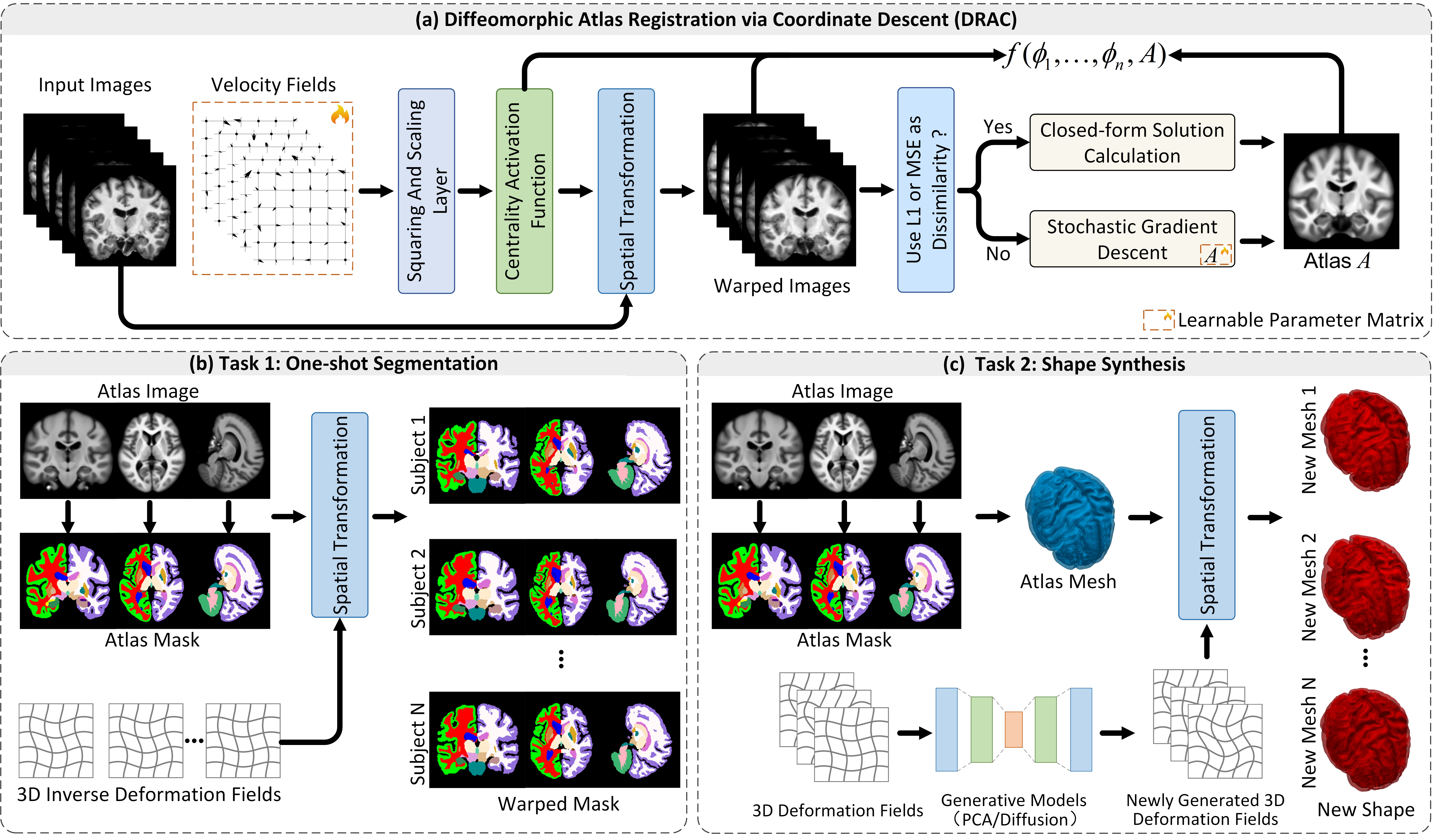}
        \vspace{-20pt}
	\caption{An overview of the proposed DARC algorithm and its two key applications: (a) DARC is a model-driven framework that jointly optimizes the velocity fields and the atlas image; (b) The application of DARC for one-shot segmentation; (c) The application of DARC for novel shape synthesis.}
        \vspace{-10pt}
	\label{Figure 1}
\end{figure*}

\vspace{5pt}
\noindent \textbf{Data-driven methods}  based on neural networks offer a modern alternative for atlas construction. AtlasMorph \cite{12} was the first to use convolutional neural networks for both unconditional and conditional atlas generation. Building on this, \cite{45} modeled attribute-dependent morphological variations as diffeomorphic transformations for condition-specific brain atlases, while \cite{14} used implicit neural representations to learn continuous spatiotemporal embeddings, enabling fetal brain atlas generation at arbitrary time points and resolutions. \cite{16} proposed an end-to-end framework that eliminates the need for affine pre-registration and introduced a novel pairwise similarity loss to improve registration accuracy. Generative models have also been applied to atlas construction. \cite{15} formulated deformable registration and conditional atlas estimation as an adversarial game, developing a GAN-based method for attribute-guided atlas generation. \cite{18} used latent diffusion models to synthesize condition-specific deformation fields from a general atlas. \cite{21} introduced InstantGroup, a VAE-based framework that performs interpolation and arithmetic in latent space to construct atlases at arbitrary scale. However, data-driven methods often require fine-tuning or retraining for different datasets or unseen subjects. As a result, their inference-time advantages over model-driven iterative approaches are not yet evident.

\section{Methodology}

Given a group of images ${\cal I}=\{ {I_i} \in \mathbb{R}^{MNH}\}_{i=1}^n $, where $M \times N \times H$ is the image size and $n$ is the number of images in the group, our goal is to construct an atlas $A \in \mathbb{R}^{MNH}$ such that all samples are mapped into a common reference space via their corresponding deformation field $\Phi=\{\phi _i \in (\mathbb{R}^{MNH})^3 \}_{i=1}^n $. This process can be formulated as a constrained optimization problem given as follows:
\begin{equation} \label{eq:1}
\mathop {\min }\limits_{{\phi _1}, \ldots, {\phi _n}, A} f(\phi_1, \ldots, \phi_n, A) \;\; s.t. \; \sum\limits_{i = 1}^n \phi_i = 0,
\end{equation}
where
\begin{equation} \label{eq:2}
f(\phi_1, \ldots, \phi_n, A) = \sum\limits_{i = 1}^n {{{\cal L}}\left( {{I_i} \circ {\phi _i}, A} \right)}  + \lambda \| {\nabla {\phi _i}} \|_2^2.   
\end{equation}
In \eqref{eq:2}, ${{\cal L}}$ indicates the dissimilarity data term, the second term is the diffusion regularization term \cite{47}, $\circ $ stands for the spatial transformation (warping), and $\lambda $ is the regularization hyperparameter. In addition, the centrality constraint $\sum \phi_i = 0$ in \eqref{eq:1} ensures that the resulting atlas image $A$ is unbiased toward any individual image in the group. To build a template or atlas for the entire dataset ${\cal I}$, both $A$ and all deformation fields $\Phi$ need to be optimized jointly in each iteration when using gradient descent. However, as the number of 3D images increases, GPU and CPU memory usage can grow exponentially, eventually exceeding the available memory.
\vspace{5pt}



\noindent 
\textbf{Coordinate descent method}: To effectively mitigate memory constraints, we introduce an asynchronous atlas update strategy based on the coordinate descent method, which is given as follows:
\begin{equation} \label{eq:cdm}
\left\{
\begin{aligned}
\phi_1^{k+1}       &= {\rm argmin}_{\phi_1} f\left(\phi_1, \phi_2^k, \ldots, \phi_n^k, A^k\right) \\
\phi_2^{k+1}       &= {\rm argmin}_{\phi_2} f\left(\phi_1^{k+1}, \phi_2, \ldots, \phi_n^k, A^k\right) \\
&\;\; \vdots \\
\phi_n^{k+1}       &= {\rm argmin}_{\phi_n} f\left(\phi_1^{k+1}, \phi_2^{k+1}, \ldots, \phi_n, A^k\right) \\
\hat{\phi}_i^{k+1} &= \phi_i^{k+1} - \frac{1}{n}\sum_{j=1}^n \phi_j^{k+1}, \quad \forall i \in \{1,\ldots,n\} \\
A^{k+1}            &= {\rm argmin}_{A} f\left(\hat{\phi}_1^{k+1}, \hat{\phi}_2^{k+1}, \ldots, \hat{\phi}_n^{k+1}, A\right)
\end{aligned}.
\right.
\end{equation}
The coordinate descent algorithm detailed in \eqref{eq:cdm} minimizes a multivariable loss function \eqref{eq:1} by optimizing one variable (or coordinate) at a time while keeping the others fixed, and iteratively cycles through each variable, updating it to reduce the loss function value until convergence.  It is also worth noting that, regardless of the data term used, the optimization over the deformation fields $\Phi=\{\phi _i \}_{i=1}^n $ can be performed in parallel, as they are not coupled variables. 

\vspace{5pt}
\noindent \textbf{Diffeomorphisms}: To ensure that the deformation fields $\Phi$ are diffeomorphic, we parameterize $\Phi$ using the stationary velocity fields ${\cal V}=\{v_i\}_{i=1}^n$ \cite{29}. Diffeomorphic deformations are differentiable and invertible, thereby preserving topology, which is essential for our one-shot segmentation and shape synthesis detailed in Section 4. In group theory, a stationary velocity field belongs to Lie algebra, and we can exponentiate it with $\phi_i=Exp({v_i})$ to obtain diffeomorphic deformations. In this paper, we employ scaling and squaring seven times \cite{29,dalca2019unsupervised} to enforce diffeomorphisms. Note that achieving diffeomorphisms requires optimizing $v_i$ in \eqref{eq:cdm} instead of directly optimizing $\phi_i$.




\vspace{5pt}
\noindent \textbf{Centrality activation function}: The penultimate equation in \eqref{eq:cdm} serves as an activation function to strictly enforce the centrality (zero-displacement) constraint, ensuring that the atlas $A$ is unbiased. This activation function is differentiable, allowing it to be used in PyTorch to leverage its automatic differentiation functionality. To this end, many methods \cite{12, 13, 15} incorporate the norm of the “average velocity field” as a regularization term during optimization and minimize it together with other energy terms. However, this introduces an additional hyperparameter that requires careful tuning, and the constraint may not be strictly satisfied.





\vspace{5pt}
\noindent \textbf{Dissimilarity metrics}: The last optimization problem in \eqref{eq:cdm} should be considered separately if the dissimilarity data term in \eqref{eq:2} is chosen differently. For this, we consider four dissimilarity metrics: mean square error (MSE), $L_1$, normalized cross-correlation (NCC) and structural similarity (SSIM). However, in principle any differentiable metric that measures the distance between the atlas image and each warped input image is compatible with the proposed algorithm. Both MSE and $L_1$ are point-wise metrics, whereas NCC and SSIM are patch-wise metrics. For MSE, the last optimization problem in \eqref{eq:cdm} reads
\begin{equation}
\begin{aligned}
    A^{k+1} = \argminB_{A} \sum\limits_{i = 1}^n {\| {{I_i} \circ {\hat{\phi}_i^{k+1}} - A} \|_2^2} = \frac{1}{n}\sum\limits_{i = 1}^n {{I_i} \circ \hat{\phi}_i^{k+1}}.
\end{aligned}
\end{equation}
In this scenario, the atlas image has a closed-form solution given by the average of all warped images at iteration $k+1$. For the $L_1$ data term, there is also a closed-form solution for the atlas image which is given by
\begin{equation}
A^{k+1} = {\rm{median}}( {I_1} \circ \hat{\phi}_1^{k+1}, {I_2} \circ \hat{\phi}_2^{k+1}, \ldots , {I_n} \circ \hat{\phi}_n^{k+1}),
\end{equation}
where ${\rm{median}}$ denotes the channel-wise median value across all warped images at iteration $k+1$. Note that with closed-form solutions, the computation is very inexpensive, so there are no out-of-memory issues for both MSE and $L_1$ data terms.

Since NCC and SSIM lack closed-form solutions, minimizing the last optimization problem with respect to the atlas image in \eqref{eq:cdm} becomes non-trivial. This is because the optimization requires all warped 3D images to be loaded into memory, which poses significant memory challenges. To overcome this, we propose using mini-batch stochastic gradient descent for the atlas update. Once the deformation fields $\Phi$ are updated and the warped images ${\cal I} \circ \Phi$ are computed, we then sample a subset (a mini-batch) of these warped images to compute gradient with respect to $A$ and perform an update on $A$. This design enables the algorithm to efficiently handle an arbitrary number of 3D images without hitting GPU memory limits. The full algorithm is given in \textbf{Algorithm 1}.  

\begin{figure}[t]
  \hrule height 1pt
  \vspace*{0.2\baselineskip}
  \centering
  \caption*{\bfseries Algorithm 1: PyTorch-style Pseudocode for DARC}
  \vspace*{-0.8\baselineskip}
  \hrule height 0.5pt
  \vspace{5pt}
  \begin{lstlisting}[xleftmargin=0pt]
def diffeo_pairwise_reg($I$,$A$,$K$,$\lambda$)
# (*@\cmtmath{I}@*): moving image; (*@\cmtmath{K}@*): number of iterations 
# (*@\cmtmath{A}@*): atlas; (*@\cmtmath{\lambda}@*): smooth parameter
    $v =$ initialize()  
    for $k$ in range($K$):                
        $\phi$ $=$ Exp($v$) # squaring and scaling
        data_term $=$ dissimilarity(${I} \circ \phi$,$A$)
        loss $=$ data_term + $\lambda \|\nabla \phi \|^2_2$
        loss.backward() # compute gradient
        optimizer.step() # optimization
    return $\phi$

def SGD_for_atlas(${\cal I}$,$\Phi$,$K$)
# (*@\cmtmath{\mathcal{I}=\{I_i\}_{i=1}^{n}}@*): all subject images;
# (*@\cmtmath{\Phi=\{\phi_i\}_{i=1}^{n}}@*): deformation fields; (*@\cmtmath{K}@*): epochs
    $A=$ initialize()
    for $k$ in range($K$):
        for ${\cal B}_{\cal I}$, ${\cal B}_{\Phi}$ in dataloader: #batch               
           loss $=$ dissimilarity(${\cal B}_{\cal I} \circ {\cal B}_{\Phi}$,$A$)
           loss.backward()
           optimizer.step()
    return $A$

def main()
    $A =$ initialize()
    for $k$ in range($K_1$):
        # step 1: compute deformations
        $\phi_i$ $=$ diffeo_pairwise_reg($I_i$,$A$,$K_2$,$\lambda$) $\forall i \in \{1,...,n\}$
        # step 2: centrality activation
        $\hat{\phi}_i = \phi_i - \frac{1}{n}\sum_j \phi_j $ $\forall i \in \{1,...,n\}$
        $\hat{\Phi} = \{\hat{\phi}_1,..., \hat{\phi}_n\}$ #collect deformations
        # step 3: update atlas  
        if dissimilarity in [$L_1$,MSE]:
            $A=$  closed_form(${\cal I} \circ \hat{\Phi}$)
        else:
            $A=$  SGD_for_atlas(${\cal I}$,$\hat{\Phi}$,$K_3$)
    return $A$,$\hat{\Phi}$
  \end{lstlisting}
  \hrule height 1pt
 \vspace{-15pt}
\end{figure}

\section{Applications of Atlas-based Modeling}
The atlas construction process produces an atlas image, along with $n$ warped images, $n$ forward deformation fields, and $n$ backward deformation fields. These results are further exploited in two novel downstream applications, namely one‑shot segmentation and shape synthesis, which were depicted in Fig \ref{Figure 1}(b) and (c), respectively.

\vspace{5pt}
\noindent \textbf{One-shot segmentation:} Segmentation of every subject image can be completed by a single segmentation mask of the atlas image and its propagation to all individual subjects via corresponding backward deformation fields. The equation used for this one-shot segmentation is presented below:
\begin{equation} \label{eq:oneshot} 
I_i^{(S)} = {A^{*(S)}} \circ Exp\left( - {v^*_i}\right),
\end{equation}
where $I_i^{(S)}$ denotes the resulting segmentation mask of the subject image ${I_i}$, while $A^{*(S)}$ denotes the segmentation mask of the optimized atlas image $A^*$. $Exp(-v^*_i)$, the inverse of $Exp(v^*_i)$, represents the deformation field that maps the atlas image to the $i$th subject image, where $v^*_i$ is the $ith$ output velocity field optimized using DARC. Note that this process requires only an annotation of the atlas image.


\vspace{5pt}
\noindent \textbf{Shape synthesis:} With the segmentation mask of the atlas image, a surface mesh consisting of vertices and triangular faces can be extracted using the marching cubes algorithm. We propose applying synthetic deformation fields to the atlas surface mesh to generate a virtual population of anatomical structures (brain and heart) with point-to-point correspondence. For this purpose, we consider two types of generative models: a simple model based on principal component analysis (PCA) and an advanced model based on diffusion processes. First, for PCA the formulation used for generating deformations is given by:
\begin{equation} \nonumber
\phi^{New} = Exp\left( { \widetilde{v}  + \sum\limits_{j = 1}^p {{\alpha _j}} {U_{\{ :{\rm{ }},{\rm{ }}j\} }}} \right), {\alpha_j}\sim{\cal N}\left(0,{{\lambda_j}}\right).
\end{equation}
Here, $\widetilde{v}$ denotes the mean velocity field computed by averaging all resulting velocity fields ${\cal V^*} = \{v_i^*\}_{i=1}^n$, and $p$ is the number of principal components (PCs) used for generation. The columns of $U$ correspond to the sorted PCs of the covariance matrix derived from ${\cal V}^*$, and $\lambda_j$ denotes the $j$th eigenvalue of the covariance matrix.

Next, the reverse equation used for generating deformation fields using the diffusion model is given by:
\begin{equation}  \nonumber
\left\{ \begin{array}{l}
{{\cal P}_\theta }(x^{0:{\rm{ }}T}) = {\cal N}({x^T};0,I)\prod\limits_{t = 1}^T {\cal N} ({x^{t - 1}};{\mu _\theta }({x^t},t),{\beta ^t}I)\\
{\phi^{New}} = Exp\left( {x^{0}} \right)
\end{array}, \right.
\end{equation} 
Where \( x \in \{v^*_1, \dots, v^*_n\} \) represents a multi-channel velocity image, \( t \in [1, \dots, T] \) denotes the diffusion timestep, and \( P_{\theta}(x^{0:T}) \) is the joint probability of the trajectory \( \{x^0, x^1, \dots, x^T\} \) under network parameters \( \theta \). The term \( \mathcal{N}(x^T; 0, I) \) denotes the Gaussian prior over \( x^T \), while \(\mathcal{N}\bigl(x^{t-1}; \mu_{\theta}(x^t, t), \beta^t I\bigr)\) represents the Gaussian distribution used to sample a slightly denoised version \(x^{t-1}\) from the current noisy input \(x^t\), with the mean predicted by a neural network and a fixed variance \(\beta^t I\).



Lastly, shape synthesis is performed by warping the atlas surface mesh using the generated (forward) deformation field $\phi^{\text{New}}$ obtained from either the PCA or diffusion model. This approach has three key novelties. First, the generated deformation fields are diffeomorphic, which prevents mesh folding and thereby ensures anatomical plausibility. Second, all generated shapes inherently preserve point-wise correspondence due to the consistent warping from the atlas. Third, the entire generation process operates in image space, without involving surface meshes, allowing direct applicability of generative models designed for images.




\section{Experiments}
\vspace{5pt}
\noindent \textbf{Datasets and implementation details:}
We evaluated atlas construction using our DARC approach on two image datasets from different organs: the brain dataset OASIS \cite{19} and the cardiac dataset \cite{20}. OASIS comprises 416 three-dimensional T1-weighted brain scans with dimensions 192 $\times$ 160 $\times$ 224. The cardiac dataset contains 220 three-dimensional cardiac MRI volumes, all of which were cropped or padded to 96 $\times$ 128 $\times$ 128. We trained DARC using the Adam optimizer with a fixed learning rate of $1\mathrm{e}{-2}$ to update both the velocity fields and atlas image across both datasets. For the cardiac dataset, a batch size of 10 was used for both velocity field estimation and stochastic gradient descent (SGD); for the OASIS dataset, a batch size of 4 was applied. For $L_1$ and MSE, the regularization parameter $\lambda = 0.5$, whereas for NCC and SSIM $\lambda = 8$. The outer iteration count $K_1=10$, and the inner iteration $K_2=300$ for updating the velocity fields. The atlas update loop $K_3=20$ when employing SGD with NCC or SSIM as the dissimilarity metric. All models were trained on a single NVIDIA RTX A6000 GPU with 48GB of RAM.

\vspace{5pt}
\noindent \textbf{Evaluation metrics:} We evaluated the accuracy of atlas construction with DARC using three metrics: Dice accuracy, deformation centrality, and the percentage of negative Jacobian determinant values (${\left| J \right|_{ < 0}}\% $), which reflects the folding rate. For one-shot segmentation, we reported the Dice score. For shape synthesis, we employed metrics like Specificity Error (mean distance from generated mesh points to original surfaces used for training) \cite{48}. Coverage (fraction of real meshes covered by generated meshes via nearest-neighbor matching), Min-Match Distance (minimum surface error under optimal point matching), Shape JSD (Jensen–Shannon divergence between the distributions of generated and real shapes), and 1-NNA (one-nearest-neighbor accuracy, where values near 50\% indicate indistinguishability and good distributional coverage) \cite{49}. Collectively, these metrics reflect the accuracy and effectiveness of the proposed DARC approach.

\begin{figure}[t]
	\centering
	\includegraphics[width=\linewidth]{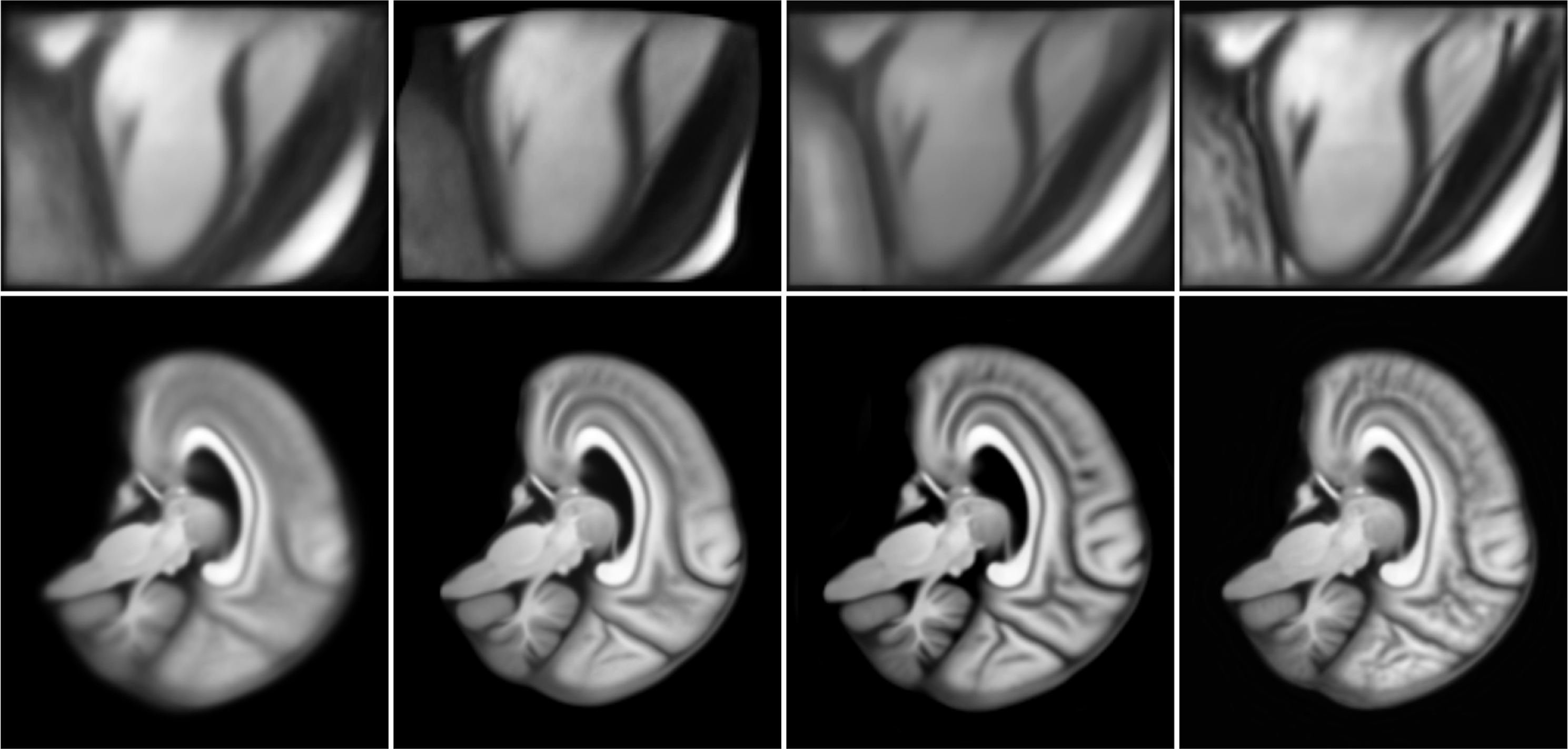}
        \vspace{-20pt}
	\caption{Visualization of atlas construction results using different dissimilarity metrics. From left to right, the results correspond to MSE, $L_1$, NCC, and SSIM, respectively.}
	\label{figure 2}
\end{figure}

\vspace{5pt}
\noindent \textbf{Impact of dissimilarity metrics:} First, we compared the performance between different dissimilarity metrics: MSE, $L_1$, NCC and SSIM. Quantitative and qualitative results were given in Table \ref{TB 1} and Fig \ref{figure 2}, respectively. From Table \ref{TB 1}, all four metrics produced folding-free atlas construction but exhibited distinct behaviors in Dice accuracy, centrality error, and qualitative performance. MSE, with its quadratic penalty, over-smooths the deformation field, but yielding the highest Dice scores on the heart. $L_1$, being median-sensitive, reduces extreme deviations but performs worst in Dice and centralization. NCC achieves the best Dice and moderate centrality error on the contrast-rich OASIS brain MRI. SSIM prioritizes structural similarity, preserving fine myocardial and cortical textures and producing the most visually detailed results, though its global Dice is slightly lower. With the centrality activation function, all methods satisfy the zero-displacement constraint, ensuring that atlas is unbiased. Fig \ref{figure 2} shows that pixel-wise metrics such as MSE and $L_1$ yield overly smoother results (with $L_1$ being the smoothest), whereas patch-wise metrics like NCC and SSIM better preserve anatomical details, with SSIM producing the most visually detailed structures. Notably, the observation that MSE performs better on the heart dataset while NCC excels on the brain dataset aligns with prior research \cite{jia2023fourier}.


\begin{table}[t]
  \centering
  \caption{Quantitative comparison of atlas construction results for different dissimilarity metrics across two datasets.}
  \vspace{-10pt}
  \resizebox{\columnwidth}{!}{%
  \begin{tabular}{cccccc}
    \toprule
    \multirow{2}{*}{\textbf{Dataset}} &
    \multirow{2}{*}{\makecell{\textbf{Metric}\\\textbf{}}} &
    \multirow{2}{*}{\makecell{\textbf{Dice }~\\ (\%, ↑)}} &
    \multirow{2}{*}{\makecell{$\boldsymbol{|J|_{<0}}\%\,$\\ (↓)}} &
    \multicolumn{2}{c}{\textbf{Centrality ($\times 10^{-3}$ ↓)}} \\
    \cmidrule(lr){5-6}
    & & & & \textbf{Before} & \textbf{After} \\
    \midrule
    \multirow{4}{*}{Heart}
      & MSE  & \textbf{86.28 ± 4.37} & 0.26 ± 0.20 & 1.2 ± 0.93 & 0.0 ± 0.0 \\
      & $L_1$   & 74.34 ± 5.09 & 0.66 ± 0.28 & 6.0 ± 6.29 & 0.0 ± 0.0 \\
      & NCC  & 74.21 ± 17.63 & 0.82 ± 2.50 & 0.9 ± 1.02 & 0.0 ± 0.0 \\
      & SSIM & 80.71 ± 5.23 & \textbf{0.0 ± 0.0} & \textbf{0.4 ± 0.46} & 0.0 ± 0.0 \\
    \midrule
    \multirow{4}{*}{OASIS}
      & MSE  & 79.54 ± 4.83 &   0.06 ± 0.04   & 0.5 ± 0.28 & 0.0 ± 0.0 \\
      & $L_1$   & 66.08 ± 5.61 &    0.02 ± 0.05   & 0.2 ± 0.13 & 0.0 ± 0.0 \\
      & NCC  & \textbf{82.44 ± 2.79} & 0.01 ± 0.01 & 0.3 ± 0.19 & 0.0 ± 0.0 \\
      & SSIM & 72.10 ± 4.53 & \textbf{0.0 ± 0.0} & \textbf{0.2 ± 0.01} & 0.0 ± 0.0 \\
    \bottomrule
  \end{tabular}}
  \label{TB 1}
\end{table}

\begin{figure}[t]
	\centering
	\includegraphics[width=\linewidth]{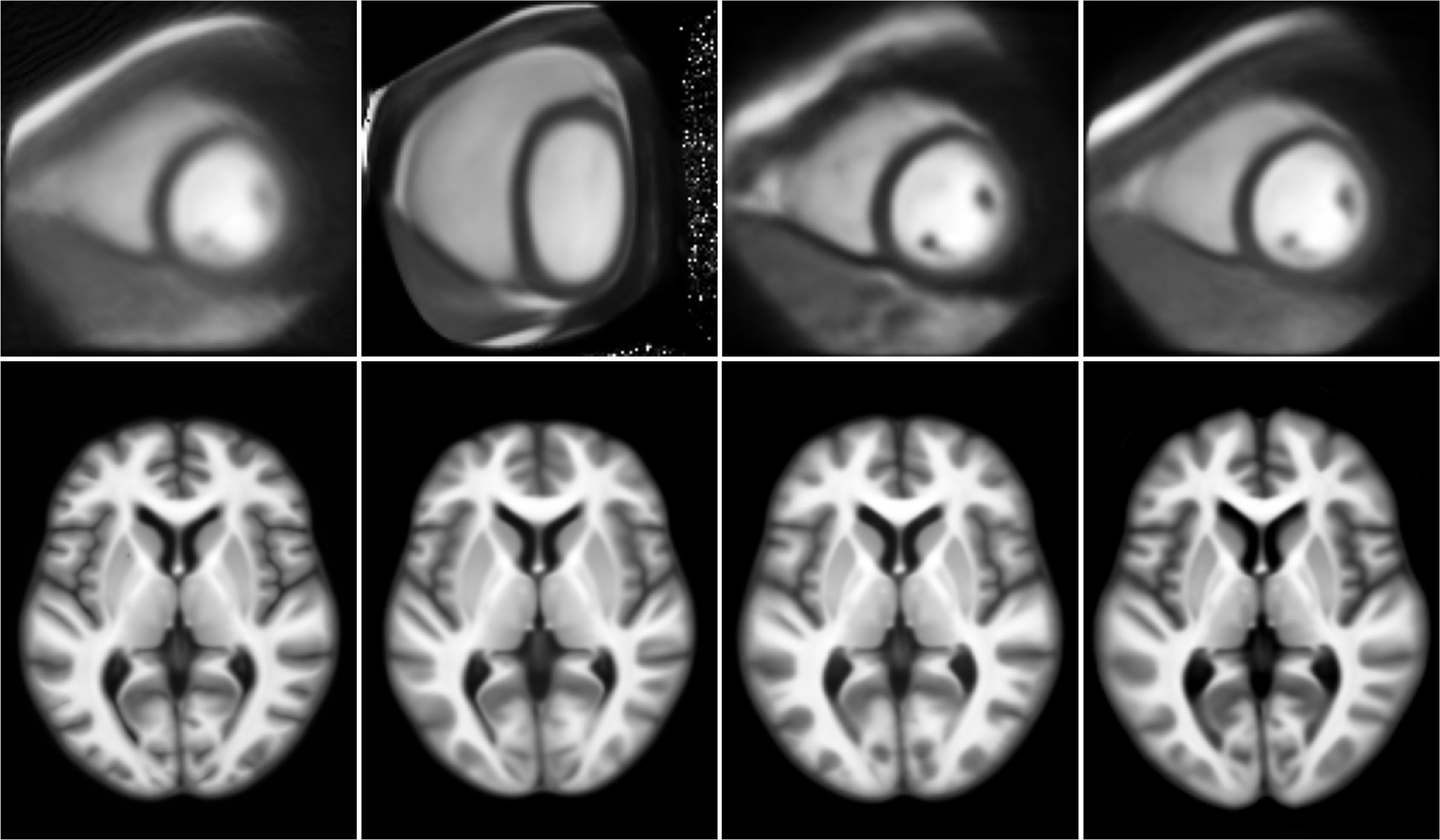}
        \vspace{-20pt}
	\caption{Visualization of atlas construction results with different methods. From left to right, the results correspond to AtlasMorph, Aladdin, MultiMorph and DARC, respectively.}
	\label{Figure 4}
\end{figure}

\begin{figure}[t]
	\centering
	\includegraphics[width=\linewidth]{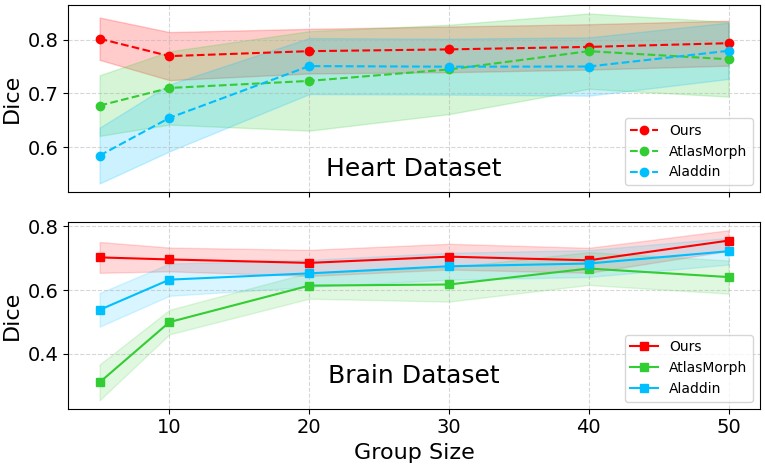}
        \vspace{-20pt}
	\caption{Performance comparison using Dice scores for atlases constructed with different numbers of subjects.}
        \vspace{-10pt}
	\label{figure 5}
\end{figure}

\begin{table}[h!]
	\centering
	\caption{Quantitative comparison of atlas construction results from state-of-the-art methods across two datasets.}
        \vspace{-10pt}
	\resizebox{\columnwidth}{!}{
		\begin{tabular}{ccccc}
			\toprule
			\textbf{Datasets} & \textbf{Method} &
			\textbf{Dice (\%, ↑)} & \makecell{${\left| J \right|_{ < 0}}\% $ (↓)} &
			\textbf{Centrality ($\times 10^{-3}$ ↓)} \\
			\midrule
			\multirow{3}{*}{Heart} 
			& AtlasMorph & $81.35 \pm 2.32$  & $0.13 \pm 0.15$  & $0.5 \pm 0.81$      \\
			& Aladdin    & $80.41 \pm 6.13$  & $\mathbf{0.0 \pm 0.0}$  & $957.0 \pm 407.84$  \\
			& Ours       & $\mathbf{86.28 \pm 4.37}$ & $0.26 \pm 0.20$  & $\mathbf{0.0 \pm 0.0}$       \\
			\midrule
			\multirow{6}{*}{OASIS} 
			& AtlasMorph & $80.27 \pm 11.03$ & $1.14 \pm 1.29$  & $0.25 \pm 0.04$     \\
			& Aladdin    & $77.39 \pm 6.46$  & $\mathbf{0.0 \pm 0.0}$  & $907.5 \pm 423.78$ \\
                & InstantGrp & $74.16 \pm 0.38$  & $\mathbf{-}$  & $168 \pm 3.0$ \\
                & ANTs & $73.55 \pm 0.44$  & $\mathbf{-}$  & $239 \pm 10.0$ \\
                & ABSORB & $73.50 \pm 0.55$  & $\mathbf{-}$  & $743 \pm 57.0$ \\
                & Ours       & $\mathbf{82.44 \pm 2.79}$ & $0.01 \pm 0.01$  & $\mathbf{0.0 \pm 0.0}$      \\
			\bottomrule
		\end{tabular}
	} 
	\label{TB 2}
\end{table}

\vspace{5pt}
\noindent \textbf{Comparison with state-of-the-art:}
We compared DARC with four recent data-driven atlas construction approaches: Multimorph \cite{22}, AtlasMorph \cite{12}, Aladdin \cite{16}, and InstantGrp \cite{21}. We also included a comparison with two model-driven methods, ANTs \cite{34} and ABSORB \cite{33}. Sourced from their GitHub repositories, we retrained the official implementations of AtlasMorph\footnote{https://github.com/voxelmorph/voxelmorph} and Aladdin\footnote{https://github.com/uncbiag/Aladdin} on the two datasets used in this paper. For AtlasMorph, we used a regularization weight of 0.1, while for Aladdin, we set the regularization weight to 10,000 and fixed the similarity loss weight at 10. A batch size of 4 was used for both methods. As Multimorph has not released full training and evaluation code, we were only able to run the inference code\footnote{https://github.com/mabulnaga/multimorph} for atlas construction with its pretrained weights. InstantGrp provides no public code, so we reported the brain results in its paper since we used a same dataset for comparison. Note that unlike the other baseline methods Multimorph incorporates mask information during training as a form of weak supervision.

Table \ref{TB 2} shows that our DRAC approach exceeds all competing methods in Dice across both datasets, achieving a folding near 0 and a centrality error of strictly 0, which evidences its optimal trade‑off between accurate alignment and topological fidelity. By imposing extremely strong regularization, Aladdin suppresses deformation field oscillations to reduce grid folding; however, its centrality error, on the order of hundreds, introduces significant global shifts away from the zero center, resulting in clear anatomical bias. Fig \ref{Figure 4} visually presents the atlases generated by each method. On the cardiac dataset, AtlasMorph produces an overly smoothed atlas, while Aladdin’s large centrality error leads to excessive deformation. In comparison, our DARC atlas demonstrates superior performance.

We further evaluated atlas construction on randomly sampled subsets of each dataset, training each model with $[5, 10, 20, \ldots, 50]$ images. As shown in Fig \ref{figure 5}, DARC consistently outperforms the other two data-driven methods, especially when the number of training images is limited. Since DARC is model-driven, its performance remains stable regardless of dataset size. Fig 1 in Appendix further confirms this trend. Overall, DARC outperforms existing methods across datasets with different anatomies, while offering a more reliable and generalizable solution, particularly in scenarios with limited data.

\begin{figure}[t]
	\centering
	\includegraphics[width=\linewidth]{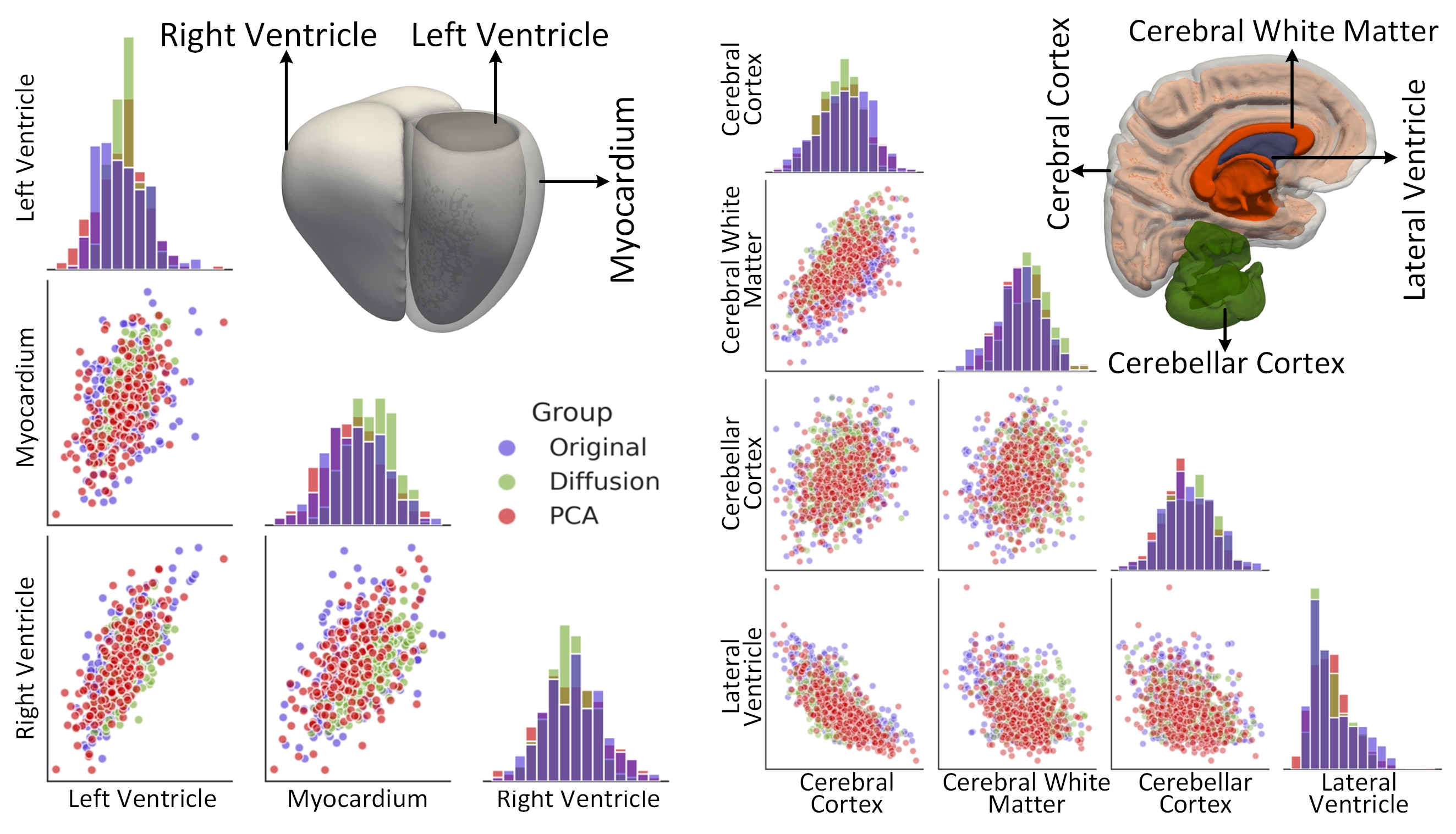}
        \vspace{-20pt}
	\caption{Visual comparison of synthetic shape distributions generated by PCA and diffusion models on heart and brain datasets. The left panel compares the distributions of the original dataset, superimposed by those generated by PCA and diffusion models, for three cardiac structures individually. The right panel presents a similar comparison for the four key regions of the left brain.}
    \vspace{-10pt}
	\label{Figure 7}
\end{figure}

\begin{table}[h!]
  \centering

  \caption{Quantitative comparison between our one-shot segmentation method and few-shot segmentation methods across two datasets.}
  \vspace{-10pt}
  \resizebox{\columnwidth}{!}{%
    \begin{tabular}{ccccc}
      \toprule
      \multirow{2}{*}{Methods} & \multicolumn{2}{c}{Heart Dice (\%, $\uparrow$)} & \multicolumn{2}{c}{OASIS Dice (\%, $\uparrow$)} \\
      \cmidrule(lr){2-3}\cmidrule(lr){4-5}
      & One-shot & Five-shot & One-shot & Five-shot \\
      \midrule
      U-Net               & $44.22\pm9.06$              & \textbf{$85.39\pm4.60$}       & $25.77\pm13.91$            & $75.14\pm8.59$            \\
      Swin UNETR              & $65.64\pm11.18$             & $82.72\pm6.14$               & \textbf{$65.66\pm8.34$}    & \textbf{$75.62\pm6.11$}   \\
      3D Attention UNet    & $52.17\pm10.31$             & $81.96\pm5.35$               & $32.67\pm12.73$            & $71.82\pm8.51$            \\
      Ours       & $\mathbf{86.28\pm4.37}$     &---                & $\mathbf{82.44\pm2.79}$                        & ---                        \\
      \bottomrule
    \end{tabular}%
  }
\label{TB 3}
\end{table}
\vspace{-10pt} 
\noindent \textbf{One-shot segmentation:}
We compared the Dice scores of our one-shot segmentation method \eqref{eq:oneshot} with those of few-shot segmentation baselines, including U-Net \cite{25}, 3D Attention U-Net \cite{24} and Swin UNETR \cite{23}. The results were reported in Table \ref{TB 3}. Our one-shot segmentation consistently outperforms these approaches in both one-shot and five-shot settings, while maintaining robust segmentation accuracy. Note that segmentation methods generally perform better on simpler anatomies, such as the heart, compared to more complex anatomies like the brain. The visual examples in Appendix Fig 2 further illustrate the advantages of our approach under limited annotation settings.

\vspace{5pt}
\noindent \textbf{Shape synthesis:} Table \ref{TB 4} shows that the diffusion model and PCA achieve similar synthesis performance on Specificity Error, Min-Match Distance, Shape JSD, and diversity/distribution alignment metrics, as further supported by their marginal histograms and scatter plots in Fig \ref{Figure 7}. While PCA effectively captures principal shape variations under its linear subspace assumption, it struggles in extreme or non-Gaussian regions, limiting its ability to represent rare morphologies and subtle nonlinear correlations. In contrast, the diffusion model better captures tail distributions and structural consistency without compromising generation accuracy, enabling more robust shape synthesis.

\begin{figure}[t]
	\centering
	\includegraphics[width=\linewidth]{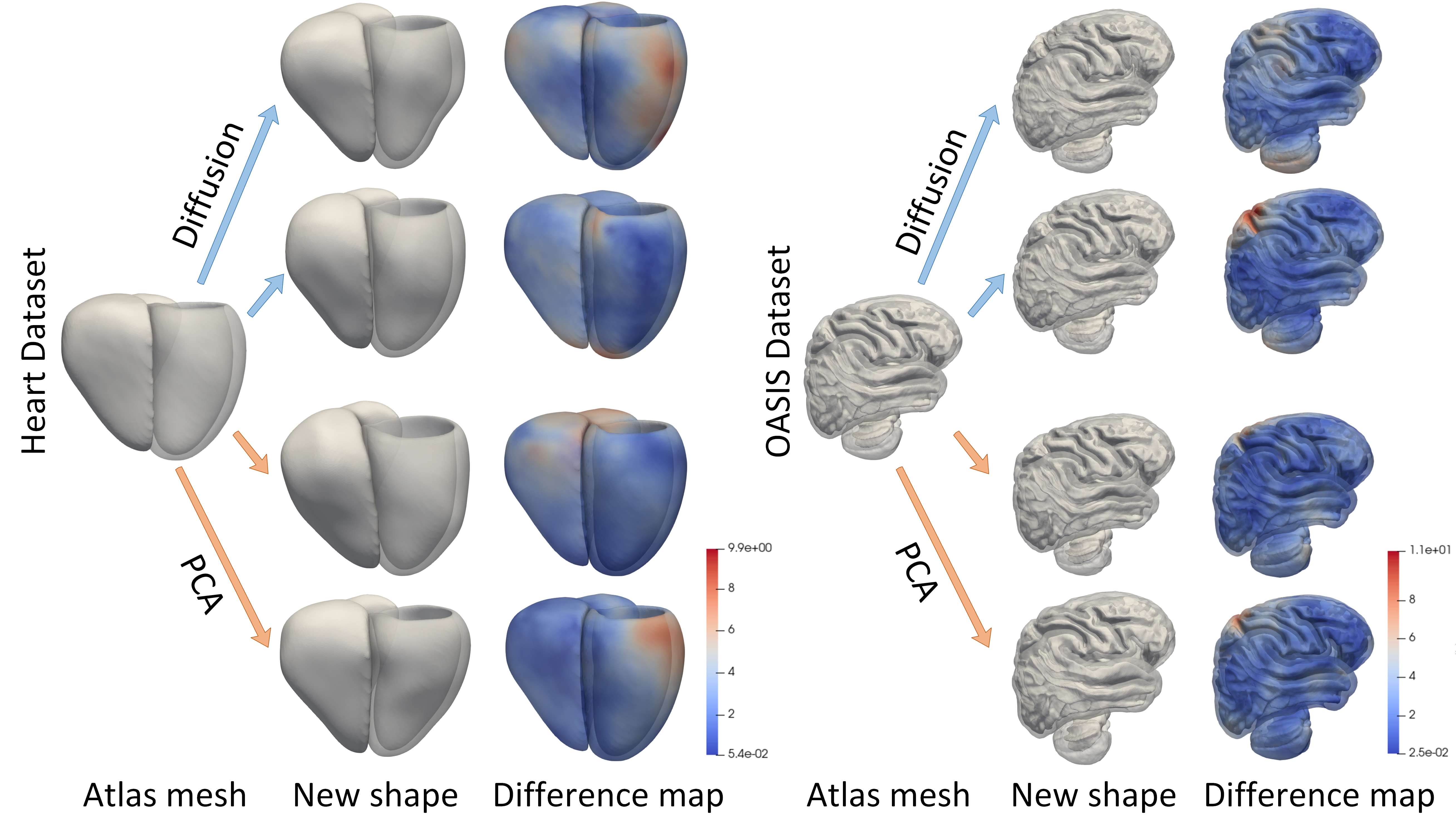}
        \vspace{-20pt}
	\caption{Visual examples of synthetic shapes of cardiac and cerebral structures generated using PCA / diffusion models.}
        \vspace{-10pt}
	\label{figure 8}
\end{figure}

In Fig \ref{figure 8}, we compared mesh samples generated by the two methods, showing that our shape synthesis approach produces diverse and topologically stable surfaces. Since all shapes were aligned with point-to-point correspondence, we performed PCA on specific anatomies to explore population-level variations. Appendix Fig 3 illustrates that, for the left ventricle, the first principal component reflects size, the second captures elongation and tilt, and the third represents elongation and roundness. For the brain, the first mode corresponds to global expansion or contraction, with uniform size changes across the whole brain; the second reveals regional variation, highlighting opposing shape changes in different cortical areas; and the third captures asymmetry and twisting, indicating localized shearing.


\vspace{-5pt}
\begin{table}[htbp]
	\centering
	\caption{Quantitative comparison of shape synthesis results using PCA and diffusion models across two datasets.}
    \vspace{-10pt}
	\resizebox{\columnwidth}{!}{%
		\begin{tabular}{ccccc}
			\toprule
			\multirow{2}{*}{\textbf{Metrics}} &
			\multicolumn{2}{c}{\textbf{Heart Dataset}} &
			\multicolumn{2}{c}{\textbf{OASIS Dataset}} \\
			\cmidrule(lr){2-3}\cmidrule(lr){4-5}
			& Diffusion & PCA & Diffusion & PCA \\
			\midrule
			Specificity Error\,\,$\downarrow$   & $\mathbf{9.43 \pm 1.35}$       & $9.59 \pm 1.51$      & $\mathbf{21.98 \pm 3.72}$      & $22.33 \pm 3.83$      \\
			Coverage\,\,( \%,\,$\uparrow$)      & $47.74$                        & $\mathbf{54.27}$      & $\mathbf{52.42}$                & $48.31$              \\
			Min-Match Distance\,\,$\downarrow$  & $9.97 \pm 2.26$                & $\mathbf{9.64 \pm 1.62}$ & $\mathbf{22.09 \pm 3.90}$       & $22.44 \pm 3.81$     \\
			Shape JSD\,\,$\downarrow$           & $0.03$                         & $\mathbf{0.01}$        & $0.01$                          & $\mathbf{0.003}$     \\
			1-NNA\,\,(50\%)                     & $64.66\,\%$           & $\mathbf{55.39\,\%}$           & $\mathbf{57.00\,\%}$            & $60.32\,\%$          \\
			\bottomrule
		\end{tabular}
	}
	\label{TB 4}
\end{table}
\vspace{-15pt} 
\section{Conclusion}
We proposed DARC, a model-driven groupwise registration approach for atlas construction. DARC incorporates a generic dissimilarity data term that supports any metric for assessing the distance between the atlas image and each warped input image. We also introduced an activation function to ensure that the resulting atlas image remains unbiased. Furthermore, we presented a one-shot segmentation framework based on the constructed atlas from DARC, reducing dependence on manually annotated images. Finally, leveraging the atlas mesh and the resulting deformation fields from DARC, we developed a novel shape synthesis method that implicitly generates novel virtual anatomical populations by synthesizing deformation fields and applying them to the atlas mesh. Experimental results on the Heart and OASIS datasets demonstrate that DARC is versatile, accurate, and unbiased. Moreover, we validated the effectiveness of the shape synthesis method and demonstrated that the proposed one-shot segmentation outperforms existing approaches in segmentation accuracy.


\bibliography{aaai2026}

\fi


\onecolumn  

\section{\Large{Appendix}}
$\;$   
\subsection{Comparison of Atlas Construction Methods}
Fig.7 compares atlas construction results using different methods and varying subject counts. Our DARC demonstrates superior stability and visualization quality compared`                   to the other two state-of-the-art methods on small-scale datasets. On the other hand, for larger-scale datasets, AtlasMorph produces overly smooth atlases, while Aladdin's substantial centering errors result in excessive deformations. In contrast, our DARC atlas demonstrates superior performance. 

\begin{figure}[htbp]
  \centering
  \includegraphics[width=\textwidth]{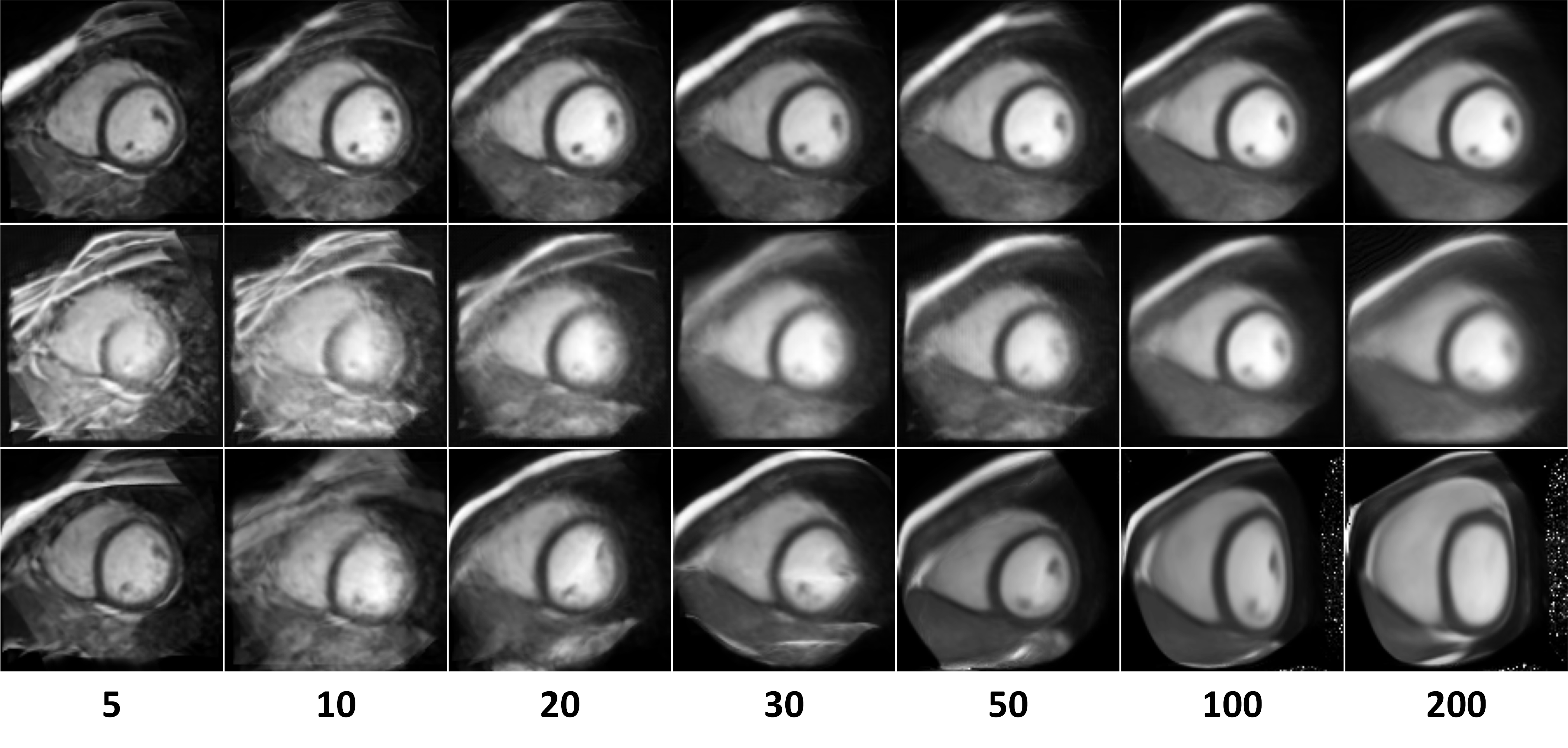}  %
  \vspace{-20px}
  \caption{Visualizations of atlases constructed by different methods using varying numbers of subjects. The rows from top to bottom represent results from DARC, AtlasMorph, and Aladdin, respectively, while the columns from left to right indicate increasing numbers of subjects.}
  \vspace{-20px}
  \label{fig1}
\end{figure}

\clearpage

\subsection{One-shot Segmentation}
Fig.8 shows one-shot segmentation results from various methods, with our DARC approach achieving the highest precision and most visually accurate results on samples from the cardiac and OASIS datasets. 
\begin{figure}[htbp]
  \centering
  \includegraphics[width=\textwidth]{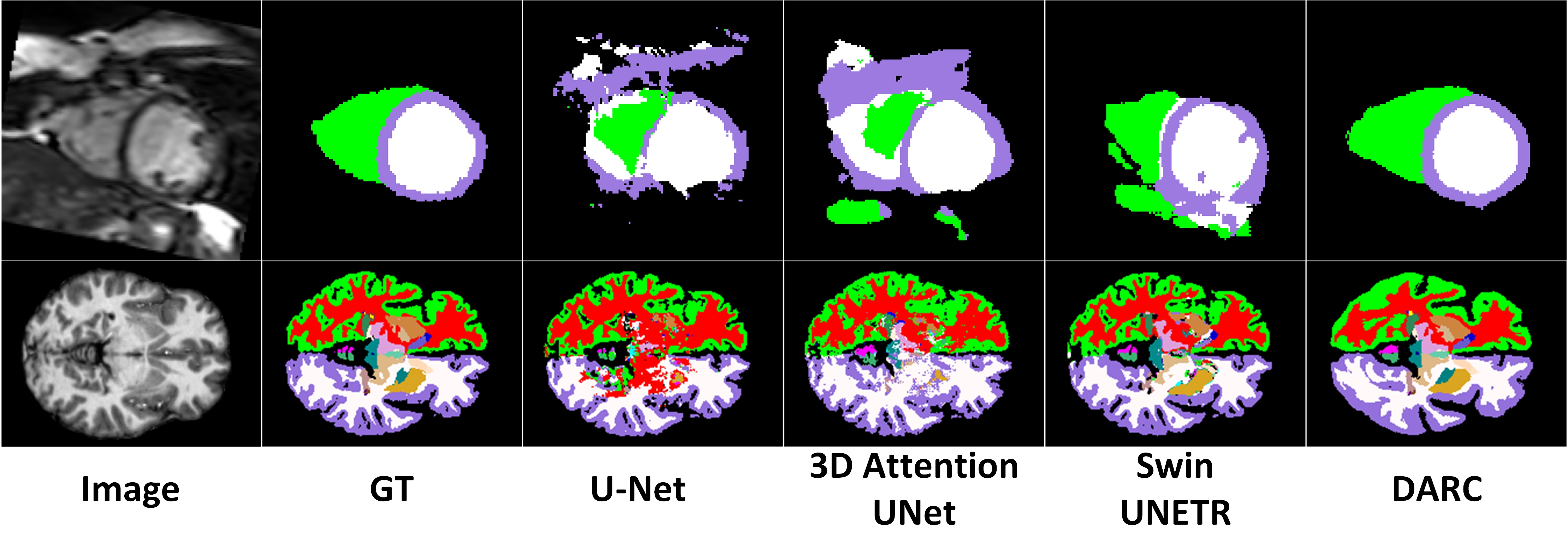} 
  \caption{Visualization of one-shot segmentation results of different methods.}
  \label{fig2}
\end{figure}

\noindent Fig.9 presents boxplots of Dice accuracy for all methods. Our DARC method achieves the highest median scores across both datasets and shows the narrowest interquartile ranges with minimal outliers, highlighting its superior accuracy and robustness.
\begin{figure}[htbp]
  \centering
  \includegraphics[width=\textwidth]{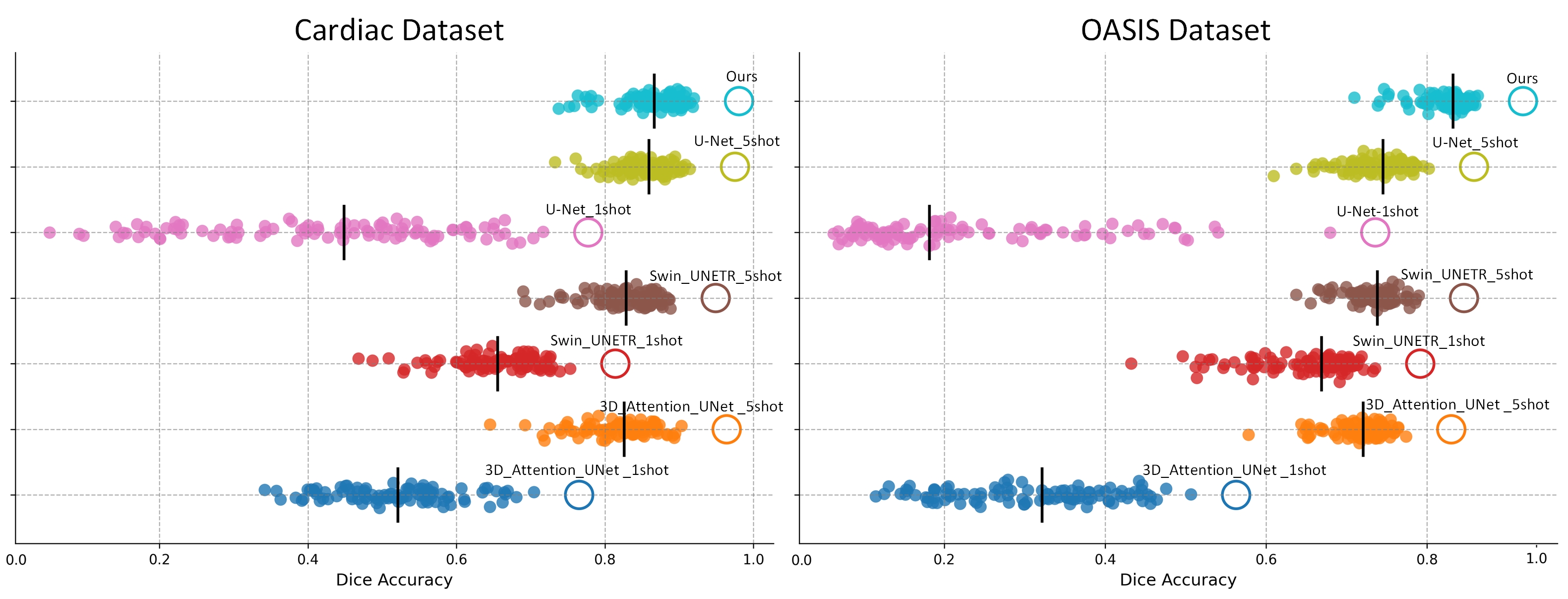}  
  \caption{A boxplot comparison of our one-shot segmentation approach against few-shot segmentation methods across two datasets.}
  \label{fig3}
\end{figure}

\clearpage

\subsection{Statistical Shape Analysis}
Fig.10 visualizes the principal component analysis (PCA) of the right cerebral white matter, the first PC (mode)  corresponds to global expansion or contraction, with uniform size changes across the whole brain; the second PC reveals regional variation, highlighting opposing shape changes in different cortical areas; and the third PC captures asymmetry and twisting, indicating localized shearing. This observation is consistent with the findings reported in \cite{63}.
\begin{figure}[htbp]
  \centering
  \includegraphics[width=\textwidth]{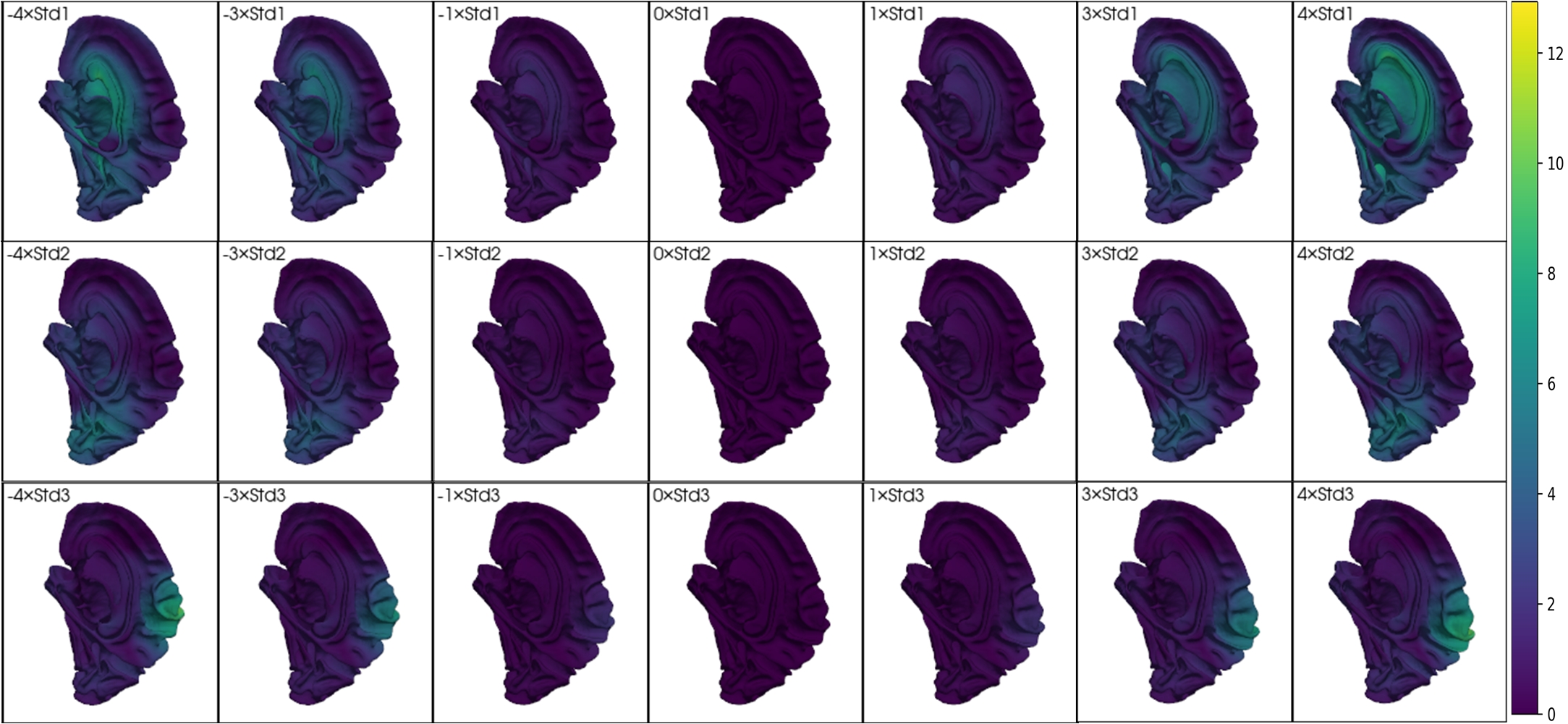}  
  \vspace{-20px}
  \caption{PCA visualization of the right cerebral white matter. Rows from top to bottom represent the first, second, and third principal components.}
  \vspace{-10px}
  \label{fig4}
\end{figure} \\

\noindent Fig.11 visualizes the principal component analysis of the left ventricle, the first PC reflects size, the second PC captures elongation and tilt, and the third PC represents elongation and roundness. This observation is consistent with the findings reported in \cite{61, 62}.\\
\begin{figure}[htbp]
  \centering
  \vspace{-10px}
  \includegraphics[width=\textwidth]{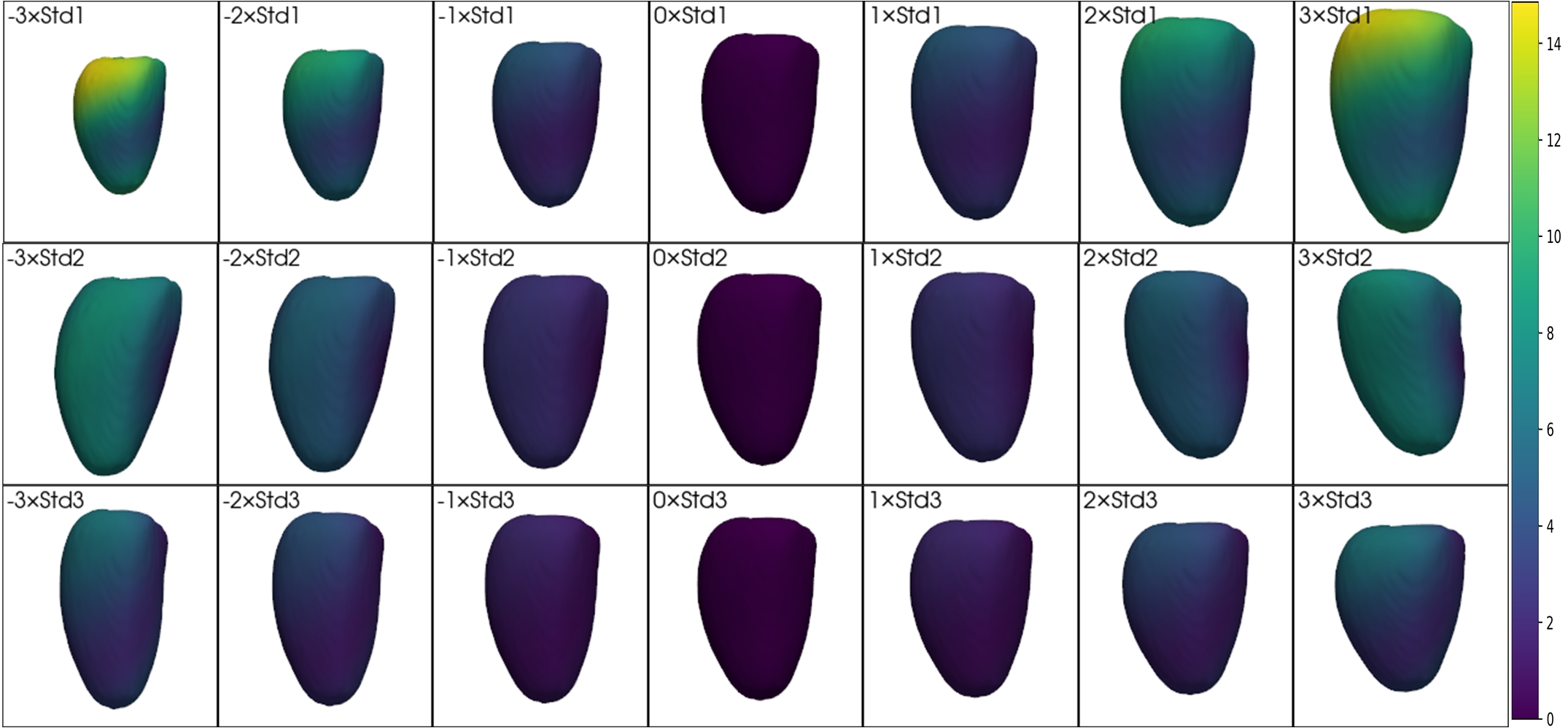}  
  \vspace{-20px}
  \caption{PCA visualization of the left ventricle, with rows from top to bottom representing the first, second, and third principal components.}
  \vspace{-20px}
  \label{fig5}
\end{figure}

\end{document}